\documentclass[letterpaper, 10 pt, journal, twoside]{IEEEtran}
\usepackage{url}
\usepackage{cite}
\usepackage{color}
\usepackage{amsmath}
\usepackage{booktabs}
\usepackage{tikz}
\usepackage{subfigure}
\usepackage{mathrsfs}
\usepackage{makecell}
\usepackage{multirow}

\usepackage[linesnumbered,ruled]{algorithm2e}
\usepackage[colorlinks=true,citecolor=black,linkcolor=black,urlcolor=black]{hyperref}

\newcolumntype{L}[1]{>{\raggedright\arraybackslash}p{#1}}
\newcolumntype{C}[1]{>{\centering\arraybackslash}p{#1}}
\newcolumntype{R}[1]{>{\raggedleft\arraybackslash}p{#1}}

\usepackage{pifont}
\newcommand{\cmark}{\ding{51}}

\newcommand{\ie}{\textit{i.e.}}
\newcommand{\eg}{\textit{e.g.}}

\newcommand{\fcn}{FCN}

\newcommand{\deeplabvthreeplus}{DeepLabv3+}

\DeclareGraphicsExtensions{.pdf,.png,.jpg,.fig}

\begin{document}
\title{Graph Attention Layer Evolves Semantic Segmentation for Road Pothole Detection:\\ A Benchmark and Algorithms}

\author{
Rui~Fan,~\IEEEmembership{Member,~IEEE},~Hengli~Wang,~\IEEEmembership{Graduate Student Member,~IEEE},~Yuan~Wang,
\\
Ming~Liu,~\IEEEmembership{Senior Member,~IEEE},~Ioannis~Pitas,~\IEEEmembership{Fellow,~IEEE}
\vspace{-1.8em}
\thanks{

\textit{Rui Fan, Hengli Wang, and Yuan Wang contributed equally to this work.}

Rui Fan is with the Department of Control Science and Engineering, College of Electronics
and Information Engineering, Tongji University, Shanghai 201804, China, and also with the Shanghai Research Institute for Intelligent
Autonomous Systems, Shanghai 201210, China (e-mail: rui.fan@ieee.org).
	
Hengli Wang and Ming Liu are with the Department of Electronic and Computer Engineering, the Hong Kong University of Science and Technology, Hong Kong SAR, China (e-mail: hwangdf@connect.ust.hk; eelium@ust.hk).
		
Yuan Wang is with the Industrial R\&D Center, SmartMore, Shenzhen 518000,
China (e-mail: yuan.wang@smartmore.com).
	
Ioannis Pitas is with the School of Informatics, University of
Thessaloniki, 541 24 Thessaloniki, Greece (e-mail: pitas@csd.auth.gr).
	}
}

\markboth{}
{Fan \MakeLowercase{\textit{et al.}}: Graph Attention Layer Evolves Semantic Segmentation for Road Pothole Detection: A Benchmark and Algorithms}

\maketitle

\begin{abstract}
Existing road pothole detection approaches can be classified as computer vision-based or machine learning-based. The former approaches typically employ 2-D image analysis/understanding or 3-D point cloud modeling and segmentation algorithms to detect (\ie, recognize and localize) road potholes from vision sensor data, \eg, RGB images and/or depth/disparity images. The latter approaches generally address road pothole detection using convolutional neural networks (CNNs) in an end-to-end manner. However, road potholes are not necessarily ubiquitous and it is challenging to prepare a large well-annotated dataset for CNN training. In this regard, while computer vision-based methods were the mainstream research trend in the past decade, machine learning-based methods were merely discussed. 
Recently, we published the first stereo vision-based road pothole detection dataset and a novel disparity transformation algorithm, whereby the damaged and undamaged road areas can be highly distinguished. However, there are no benchmarks currently available for state-of-the-art (SoTA) CNNs trained using either disparity images or transformed disparity images. Therefore, in this paper, we first discuss the SoTA CNNs designed for semantic segmentation and evaluate their performance for road pothole detection with extensive experiments. Additionally, inspired by graph neural network (GNN), we propose a novel CNN layer, referred to as graph attention layer (GAL), which can be easily deployed in any existing CNN to optimize image feature representations for semantic segmentation. Our experiments compare GAL-DeepLabv3+, our best-performing implementation, with nine SoTA CNNs on three modalities of training data: RGB images, disparity images, and transformed disparity images. The experimental results suggest that our proposed GAL-DeepLabv3+ achieves the best overall pothole detection accuracy on all training data modalities. The source code, dataset, and benchmark are publicly available at \url{mias.group/GAL-Pothole-Detection}.   
\end{abstract}

\section{Introduction}
\label{sec.introduction}

\IEEEPARstart{A} pothole is a large structural road failure \cite{mathavan2015review}. Its formation is due to the combined presence of water and traffic \cite{fan2021rethinking}. Water permeates the ground and weakens the soil under the road surface while traffic subsequently breaks the affected road surface, resulting in the removal of road surface chunks \cite{fan2019pothole}. Road potholes, besides being an inconvenience, are also a safety hazard because they can severely affect driving comfort, vehicle condition, and traffic safety \cite{fan2019pothole}. Therefore, frequently inspecting and repairing road potholes is a crucial road maintenance task \cite{mathavan2015review}. Currently, road potholes are regularly detected and reported by certified inspectors \cite{fan2020we}. This manual visual inspection process is tedious, dangerous, costly, and time-consuming \cite{fan2019pothole}. Moreover, manual road pothole detection results are qualitative and subjective as they depend entirely on individual inspectors' experience \cite{koch2015review}. Consequently, there is an ever-increasing need for automated road pothole detection systems, especially ones developed based on state-of-the-art (SoTA) computer vision and machine learning techniques. 

\begin{figure}[!t]
	\centering
	\includegraphics[width=0.43\textwidth]{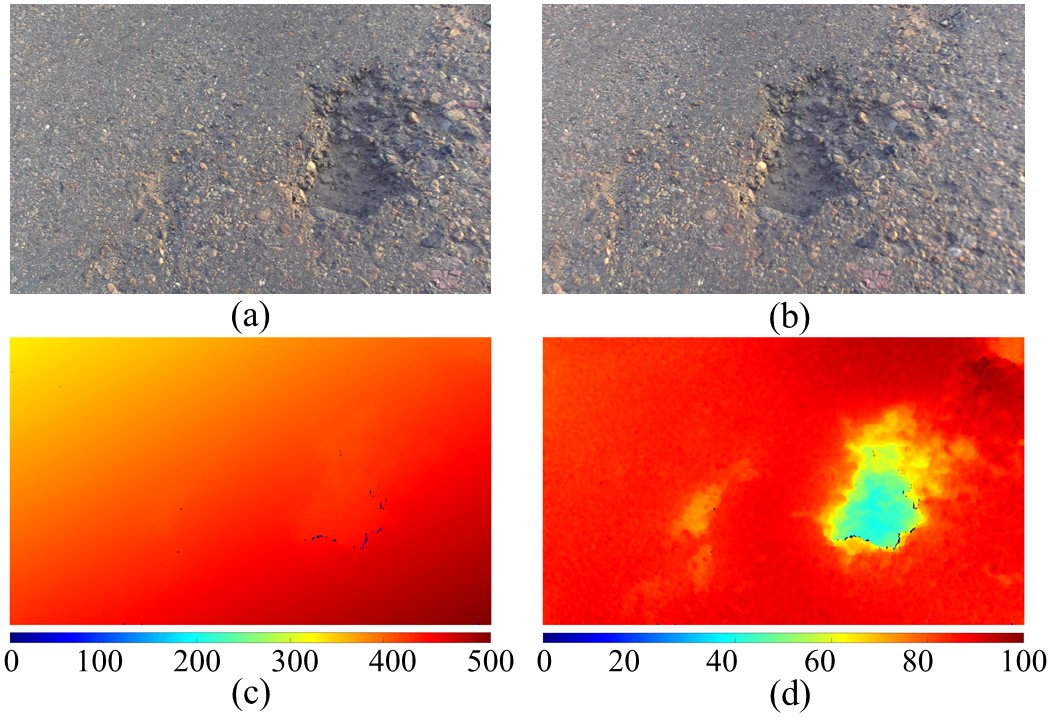}
	\caption{An example of the multi-modal road vision data provided in \cite{fan2019pothole}: (a) left road image; (b) right road image; (c) dense subpixel disparity image estimated from (a) and (b) using the stereo matching algorithm presented in \cite{fan2018road}; (d) transformed disparity image yielded from (c) using the disparity transformation algorithm introduced in \cite{fan2019pothole}.}
	\label{fig.disp_estimation_transformation}
	\vspace{-1.5em}
\end{figure}

In \cite{fan2019pothole}, we published the world's first multi-modal road pothole detection dataset, containing RGB images, subpixel disparity images, and transformed disparity images. An example of these three modalities of road vision data is shown in Fig. \ref{fig.disp_estimation_transformation}. It can be observed that the damaged road areas are highly distinguishable after disparity transformation, making road pothole detection much easier. However, there lacks a benchmark for road pothole detection based on SoTA semantic segmentation convolutional neural networks (CNNs), trained on spatial vision data, \eg, disparity/depth images, other than road RGB images. Therefore, there is a strong motivation to provide a comprehensive comparison for SoTA CNNs w.r.t. different modalities of road vision data. Additionally, some semantic segmentation approaches  \cite{krahenbuhl2011efficient, zheng2015conditional} combine CNNs with graph models, such as conditional random fields (CRFs), to improve the image segmentation performance. Such CRF-based approaches are nevertheless very computationally intensive, and thus they can only be deployed on the final semantic probability map. Therefore, a graph attention layer that can produce additional weights based on the relational inductive bias to refine image feature representations is also a research topic that requires more attention.  

We introduce a novel graph model-based semantic segmentation CNN for road pothole detection, of which the effectiveness is demonstrated with extensive experiments. The contributions of this paper are summarized as follows:
\begin{itemize}[noitemsep]
\item A benchmark of all SoTA semantic segmentation CNNs trained on the three aforementioned modalities of vision data for road pothole detection;
\item Graph attention layer (GAL), a novel graph layer inspired by graph neural network (GNN) \cite{Graph}, which can be easily deployed in any CNN to optimize image feature representations for semantic image segmentation;
\item A novel CNN, referred to as GAL-DeepLabv3+, incorporating the proposed GAL in DeepLabv3+ \cite{chen2018encoder}. It outperforms SoTA CNNs for road pothole detection.
\end{itemize}

The remainder of this paper is structured as follows: Section \ref{sec.related_work} reviews the SoTA road pothole detection approaches and semantic segmentation CNNs. Section \ref{sec.methodology} introduces our proposed methodology. In Section \ref{sec.experimental_results_and_discussion}, we compare the performance of our proposed method with the SoTA CNNs reviewed in Section \ref{sec.related_work}. Section \ref{sec.discussion} discusses the implications and practical application of this work. Finally, Section \ref{sec.conclusion_future_work} summarizes the paper and provides recommendations for future work.

\section{Literature Review}
\label{sec.related_work}

\subsection{Road Pothole Detection}
\label{sec.road_pothole_detection}
The existing road pothole detection methods can be categorized into two groups: explicit programming-based \cite{fan2019pothole} and machine learning-based \cite{fan2020we}. 

Explicit programming-based methods detect potholes via either two-dimensional (2-D) image analysis/understanding or three-dimensional (3-D) road point cloud modeling and segmentation \cite{fan2019pothole}. As an example, a Microsoft Kinect sensor is used to capture road depth images \cite{jahanshahi2012unsupervised}, which are then segmented using image thresholding method for road pothole detection. In order to ensure the applicability of image thresholding method, the Microsoft Kinect sensor has to be mounted as perpendicularly as possible to the road surface as it requires uniformly distributed background (undamaged road areas) depth values \cite{jahanshahi2012unsupervised}. The 3-D road point cloud modeling and segmentation approaches \cite{ozgunalp2016vision} typically interpolate a road surface point cloud into an explicit mathematical model, \eg, a quadratic surface. The road potholes can then be effectively detected by comparing the difference between and actual and the interpolated 3-D road surfaces. Recently, \cite{fan2019pothole} introduced a hybrid road pothole detection system developed based on disparity transformation and modeling. The disparity transformation algorithm can not only estimate the stereo camera's roll and pitch angles but also transform the disparity image into a quasi-inverse perspective view, where the background disparity values become very similar \cite{wang2021dynamic}. This algorithm does not require the depth sensor's optical axis to be perpendicular to the road surface, greatly enhancing the robustness and adaptability of depth/disparity image segmentation algorithms. Subsequently, a quadratic surface is fitted to the disparities in the undamaged road regions for accurate road pothole detection. 

Machine learning-based methods generally train CNNs on well-annotated vision data for end-to-end road pothole detection \cite{fan2019road}. Any general-purpose semantic/instance segmentation CNN can be easily applied to detect road potholes from RGB or disparity/depth images. For example, mask region-based CNN (R-CNN) is employed in \cite{dhiman2019pothole} to detect road potholes from RGB images. In \cite{wu2019road}, \deeplabvthreeplus \cite{chen2018encoder} is utilized to segment RGB images for road pothole detection. In \cite{fan2020we}, five single-modal and three data-fusion CNNs are compared in terms of detecting road potholes from RGB and/or transformed disparity images, where an attention aggregation framework and an adversarial domain adaptation technique are used to boost the CNN performance. 

\subsection{Semantic Segmentation Networks}
\label{sec.semantic_segmentation}

A fully convolutional network (\fcn) \cite{long2015fully} is an end-to-end, pixel-to-pixel semantic segmentation network. It converts all fully connected (FC) layers to convolutions. An {\fcn} typically consists of a downsampling path and an upsampling path. The downsampling path employs a classification network as the backbone to capture semantic/contextual information, while the upsampling path fully recovers the spatial information using skip connections. \fcn-32s, \fcn-16s, and \fcn-8s are three main variants having different upsampling strides to provide coarse, medium-grain, and fine-grain semantic image segmentation results, respectively. In the paper, \fcn-8s is used for road pothole detection. 

U-Net \cite{ronneberger2015u} was designed based on FCN. It was extended to work with fewer training samples while yielding more accurate segmentation results. U-Net consists of a contracting path and an expansive path. The contracting path has a typical CNN architecture of convolutions, rectified linear units (ReLUs), and max pooling layers. At the same time, the expansive path combines the desired visual features and spatial information through a sequence of upconvolutions and concatenations. The skip connection between the contracting path and the expansive path helps restore small objects' locations better. Compared with FCN, U-Net has a large number of feature channels in upsampling layers, allowing it to propagate context information to the layers with higher resolution.

SegNet \cite{badrinarayanan2017segnet} has an encoder-decoder architecture. The encoder network employs VGG-16 \cite{simonyan2014very} to generate high-level feature maps, while the decoder network upsamples its input to produce a sparse feature map, which is then fed to a softmax classifier for pixel-wise classification. Therefore, SegNet has a trainable decoder filter bank, while an FCN does not. The network depth $k$ determines the image downsampling and upsampling by an overall factor of $2^k \times 2^k$.

DeepLabv3+ \cite{chen2018encoder} is developed based on DeepLabv1 \cite{chen2014semantic}, DeepLabv2 \cite{chen2017deeplab} and DeepLabv3 \cite{chen2017rethinking}. It combines the advantages of both the spatial pyramid pooling (SPP) module and the encoder-decoder architecture. Compared to DeepLabv3, it adds a simple yet efficient decoder module to refine the semantic segmentation. Additionally, it employs depth-wise separable convolution to both atrous SPP (ASPP) and decoder modules, making its encoder-decoder structure much faster.

Although ASPP can generate feature maps by concatenating multiple atrous-convolved features, the resolution of these maps is typically not dense enough for applications requiring high accuracy \cite{chen2017deeplab}. In this regard, DenseASPP \cite{yang2018denseaspp} was developed to connect atrous convolutional layers (ACLs) more densely. The ACLs are organized in a cascade fashion, where the dilation rate increases layer by layer. Then, DenseASPP concatenates the output from each atrous layer with the input feature map and the outputs from lower layers. The concatenated feature map is then fed into the following layer. DenseASPP's final output is a feature map generated by multi-scale and multi-rate atrous convolutions. DenseASPP is capable of generating multi-scale features that cover a larger and denser scale range without significantly increasing the model size.

Different from the CNNs mentioned above, the pyramid attention network (PAN) \cite{li2018pyramid} combines an attention mechanism and a spatial pyramid to extract accurate visual features for semantic segmentation. It consists of a feature pyramid attention (FPA) module and a global attention upsample (GAU) module. The FPA module encodes the spatial pyramid attention structure on the high-level output and combines global pooling to learn better feature representations. The GAU module provides global context as a guidance for using low-level visual features to select category localization details.

\begin{figure*}[!t]
	\centering
	\includegraphics[width=0.85\textwidth]{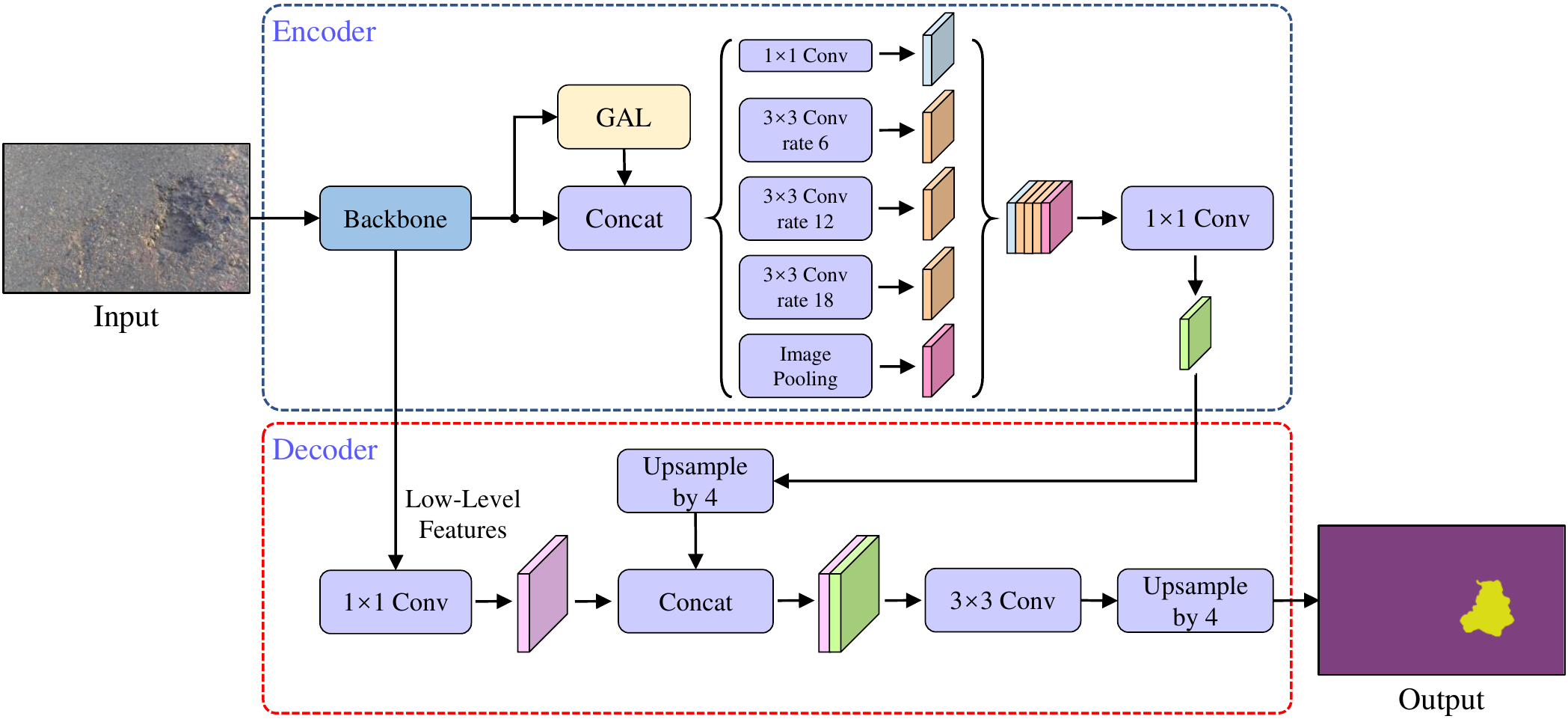}
	\centering
	\caption{The overview of our proposed GAL-DeepLabv3+. GAL takes the feature from the backbone as the input and it outputs the refined feature, which is then concatenated with the input feature and sent to the following ASPP module and the decoder.}
	\label{fig.proposed_cnn}
	\vspace{-0.5em}
\end{figure*}

\begin{figure*}[!t]
	\centering
	\includegraphics[width=0.75\textwidth]{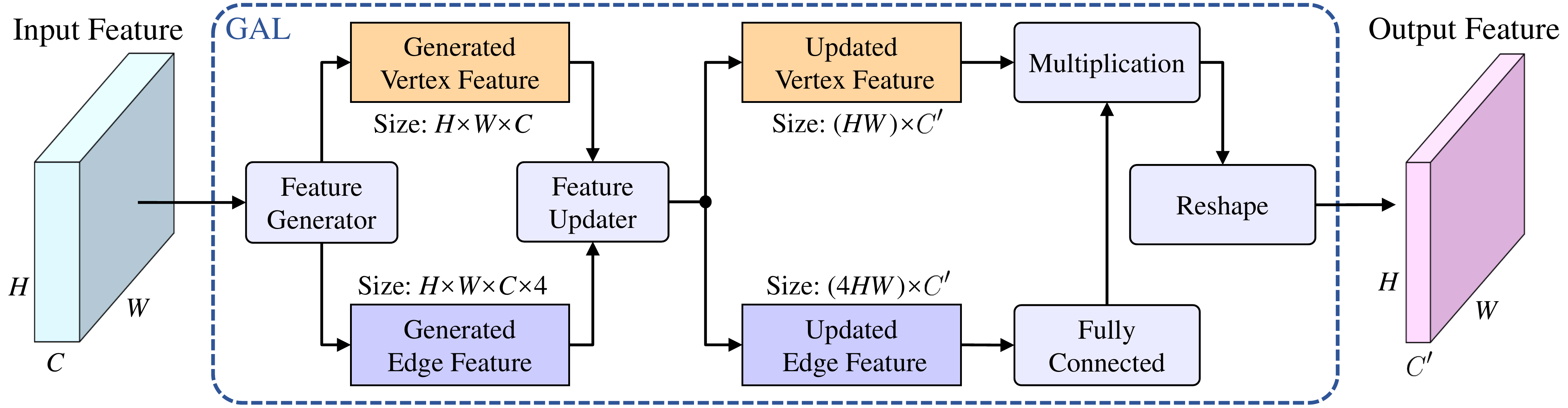}
	\centering
	\caption{An illustration of our proposed GAL, which utilizes a feature generator and a feature updater to optimize the input feature of size $H \times W \times C$ and output the refined feature of size $H \times W \times C'$. }
	\label{fig.gal}
	\vspace{-1.5em}
\end{figure*}

As discussed above, the recent CNNs with encoder-decoder architectures typically perform bilinear upsampling in the last decoder layer for final pixel-wise region prediction. However, simple bilinear upsampling limits the recovered pixel-wise prediction accuracy because it does not take pixel prediction correlations into account \cite{tian2019decoders}. DUpsampling was therefore introduced to recover the pixel-wise prediction from low-resolution CNN outputs by exploiting the redundancy in the semantic segmentation label space. It allows the decoder to downsample the fused visual features to the lowest feature map resolution before merging them. This approach not only reduces the decoder computation cost but also decouples fused feature resolution from the final prediction.

Although most deep CNNs have achieved compelling semantic segmentation results, large networks are generally slow and computationally intensive. ESPNet \cite{mehta2018espnet} was designed to resolve this issue. An efficient spatial pyramid (ESP) module consists of point-wise convolutions to help reduce the computational
complexity, and a spatial pyramid of dilated convolutions to resample the feature maps to learn the representations from the large effective receptive field. The ESP module's large effective field introduces gridding artifacts, which are then removed using hierarchical feature fusion. A skip-connection between the input and output is also added to improve the information flow.

Unlike the above-mentioned CNNs, gated shape CNN (GSCNN) \cite{takikawa2019gated} leverages a two-branch architecture. The regular branch can be any backbone CNN, while the shape branch processes shape information in parallel to the regular branch through a set of residual blocks and gated convolutional layers. Furthermore, GSCNN uses higher-level activations in the regular branch to gate the lower-level activations in the shape branch, effectively reducing noise and helping the shape branch to focus only on the relevant boundary information. This, in turn, efficiently improves the performance of the regular branch. GSCNN then employs an ASPP to combine the information from the two branches in a multi-scale fashion. The experimental results demonstrate that this architecture can produce sharper predictions around region boundaries and can significantly boost the semantic segmentation accuracy for thinner and smaller objects.

Graph models can produce useful representations for pixel relations, which greatly helps to improve the semantic segmentation performance. As discussed in Section \ref{sec.introduction}, the CRF-based approaches are computationally intensive, and can only be deployed on the final semantic probability map. To solve this disadvantage, we propose GAL, which is capable of producing additional weights based on the relational inductive bias to refine image feature representations in a computationally efficient manner.

\section{Methodology}
\label{sec.methodology}

The architecture of our introduced semantic segmentation network for road pothole detection is shown in Fig. \ref{fig.proposed_cnn}.  An initial feature is learned from the input image by the backbone CNN. It is then fed into our proposed GAL to produce a refined feature, which is concatenated with the input feature and sent to the following ASPP module.

\subsection{Graph Attention Layer}
\label{sec.gal}
A graph is commonly defined as a three-tuple $\mathscr{G}(\mathbf{u},\mathscr{V},\mathscr{E})$, where $\mathbf{u}$ is a global attribute, $\mathscr{V}=\{\mathbf{v}_k\}_{k=1:N^v}$ is the vertex set, and $\mathscr{E}=\{(\mathbf{e}_k,r_k,s_k)\}_{k=1:N^e}$ is the edge set ($r_k$ and $s_k$ represents the index of the receiver and sender vertex, respectively) \cite{Graph}.

As illustrated in Fig. \ref{fig.gal}, our proposed GAL consists of two main components: a feature generator, which generates the representations of both vertex and edge features, and a feature updater which updates these two types of feature representations based on our proposed GNN block. Then, we implement our GAL in {\deeplabvthreeplus} \cite{chen2018encoder} and refer to it as GAL-DeepLabv3+, as illustrated in Fig. \ref{fig.proposed_cnn}. The remaining subsections detail the feature generator and updater of our GAL, as well as our GAL-DeepLabv3+, separately.

\subsubsection{GAL Feature Generator}
\label{sec.feature_generator}
\begin{figure}[!t]
	\centering
	\includegraphics[width=0.36\textwidth]{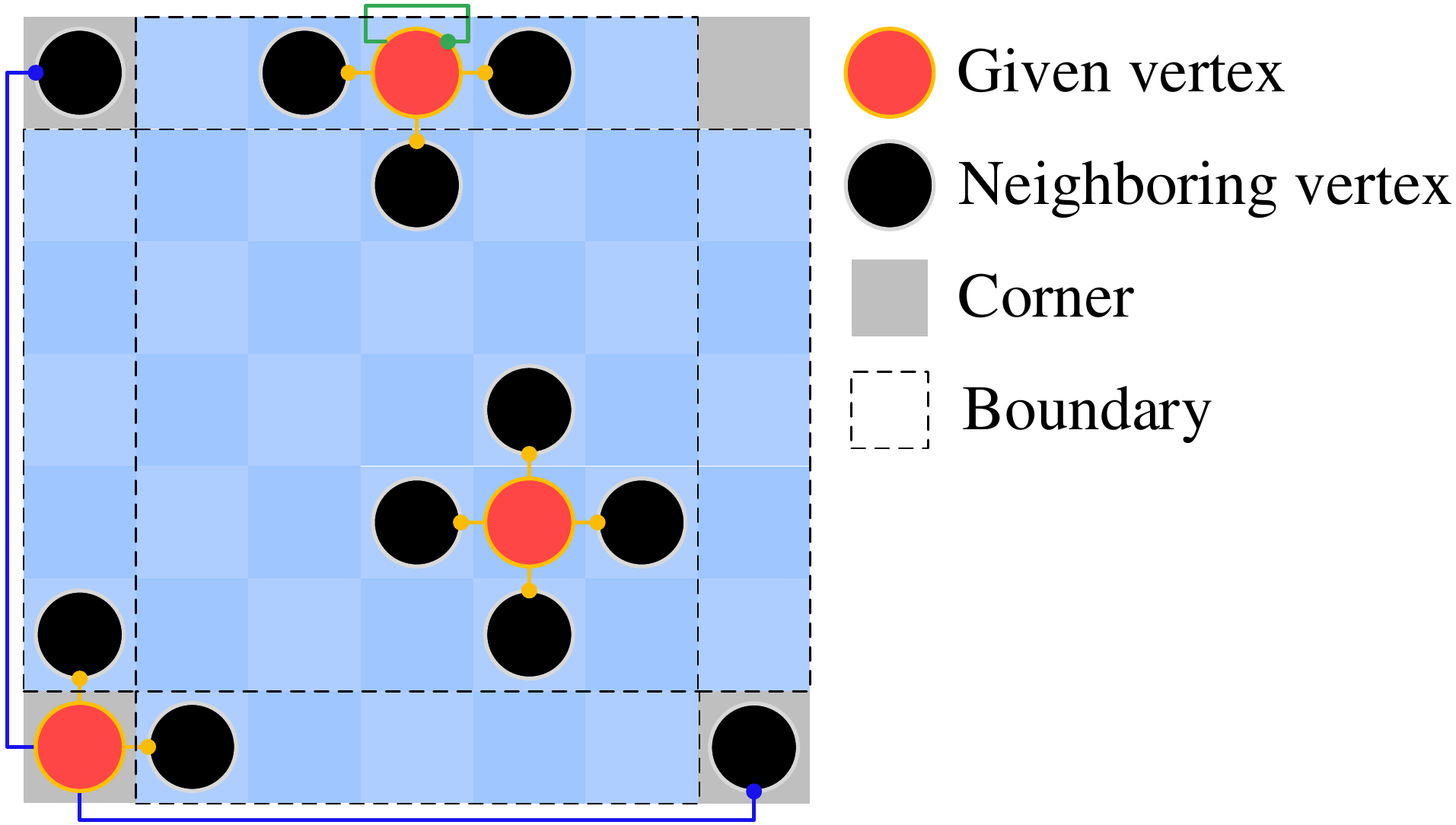}
	\centering
	\caption{An illustration of the edge feature generation, where only four closest neighbors are considered for each vertex. Moreover, when the vertex locates at a corner or on a boundary, the corresponding special neighbors are marked with blue and green lines, respectively.}
	\label{fig.feature_generation}
	\vspace{-1.5em}
\end{figure}

As shown in Fig. \ref{fig.gal}, the input of our GAL is a tensor (feature representation) $T$ of size $H \times W \times C$. $T$ is first converted to a graph $\mathscr{G}(\mathbf{u},\mathscr{V},\mathscr{E})$, where $\mathscr{V}=\{\mathbf{v}_k\}_{k=1:HWC}$ and $\mathscr{E}=\{(\mathbf{e}_k,r_k,s_k)\}_{k=1:4HWC}$ (only four closest neighbors are considered for each vertex). These are then considered as the vertex and edge features, respectively. An illustration of the edge feature generation is shown in Fig.~\ref{fig.feature_generation}, where it can be seen that there exist two special cases: corners and boundaries. When a given vertex is at the graph corner, the vertexes at another two corners are considered to be its neighbors. Moreover, when a given vertex is on the graph boundary, the vertex itself will be considered as one of its four neighbors. The generated vertex and edge feature will then be updated using a simplified GNN block.

\subsubsection{GAL Feature Updater}
\label{sec.feature_updater}
\begin{figure}[!t]
	\centering
	\includegraphics[width=0.4\textwidth]{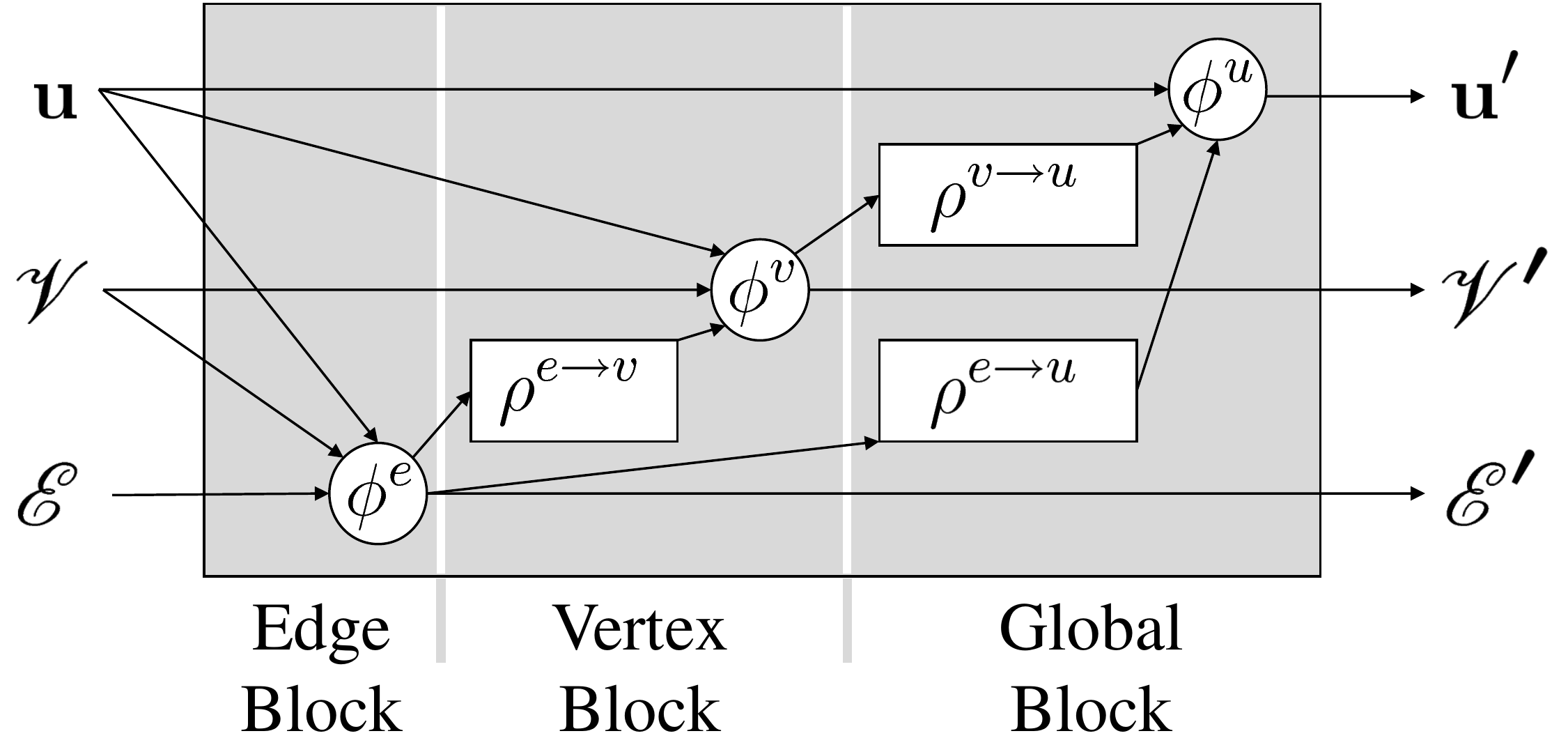}
	\centering
	\caption{An illustration of a full GNN block, which consists of an edge block, a vertex block and a global block.}
	\vspace{-1.5em}
	\label{fig.full_gnn_block}
\end{figure}

A general full GNN block is illustrated in Fig. \ref{fig.full_gnn_block}. It can be seen that it consists of three sub-blocks: an edge block, a vertex block, and a global block. Each full GNN block also contains three update functions $\phi^{e}$, $\phi^{v}$, and $\phi^{u}$ of the following forms \cite{Graph}:
\begin{equation}
\mathbf{e}'_k = \phi^{e}(\mathbf{e}_k, \mathbf{v}_{r_k}, \mathbf{v}_{s_k}, \mathbf{u}),
\label{eq.phi^e}
\end{equation}
\begin{equation}
\mathbf{v}'_i = \phi^{v}(\bar{\mathbf{e}}_i', \mathbf{v}_{i}, \mathbf{u}),
\label{eq.phi^v}
\end{equation}
\begin{equation}
\mathbf{u}' = \phi^{u}(\bar{\mathbf{e}}', \bar{\mathbf{v}}', \mathbf{u}),
\label{eq.phi^u}
\end{equation}
and three aggregation functions $\rho^{e\rightarrow v}$, $\rho^{e\rightarrow u}$, and $\rho^{v\rightarrow u}$ of the following forms \cite{Graph}:
\begin{equation}
\bar{\mathbf{e}}'_i = \rho^{e\rightarrow v}(\mathscr{E}'_i),
\label{eq.rho^e_v}
\end{equation}
\begin{equation}
\bar{\mathbf{e}}' = \rho^{e\rightarrow u}(\mathscr{E}'),
\label{eq.rho^e_u}
\end{equation}
\begin{equation}
\bar{\mathbf{v}}' = \rho^{v\rightarrow u}(\mathscr{V}'),
\label{eq.rho^v_u}
\end{equation}
where $\mathscr{E}'_i=\{(\mathbf{e}'_k,r_k,s_k)\}_{r_k=i, k= 1:N^e}$, $\mathscr{V}'=\{\mathbf{v}'_i\}_{i=1:N^v}$, and $\mathscr{E}'=\bigcup_i \mathscr{E}_{i}'=\{(\mathbf{e}'_k,r_k,s_k)\}_{k= 1:N^e}$.

\begin{figure}[!t]
	\centering
	\includegraphics[width=0.4\textwidth]{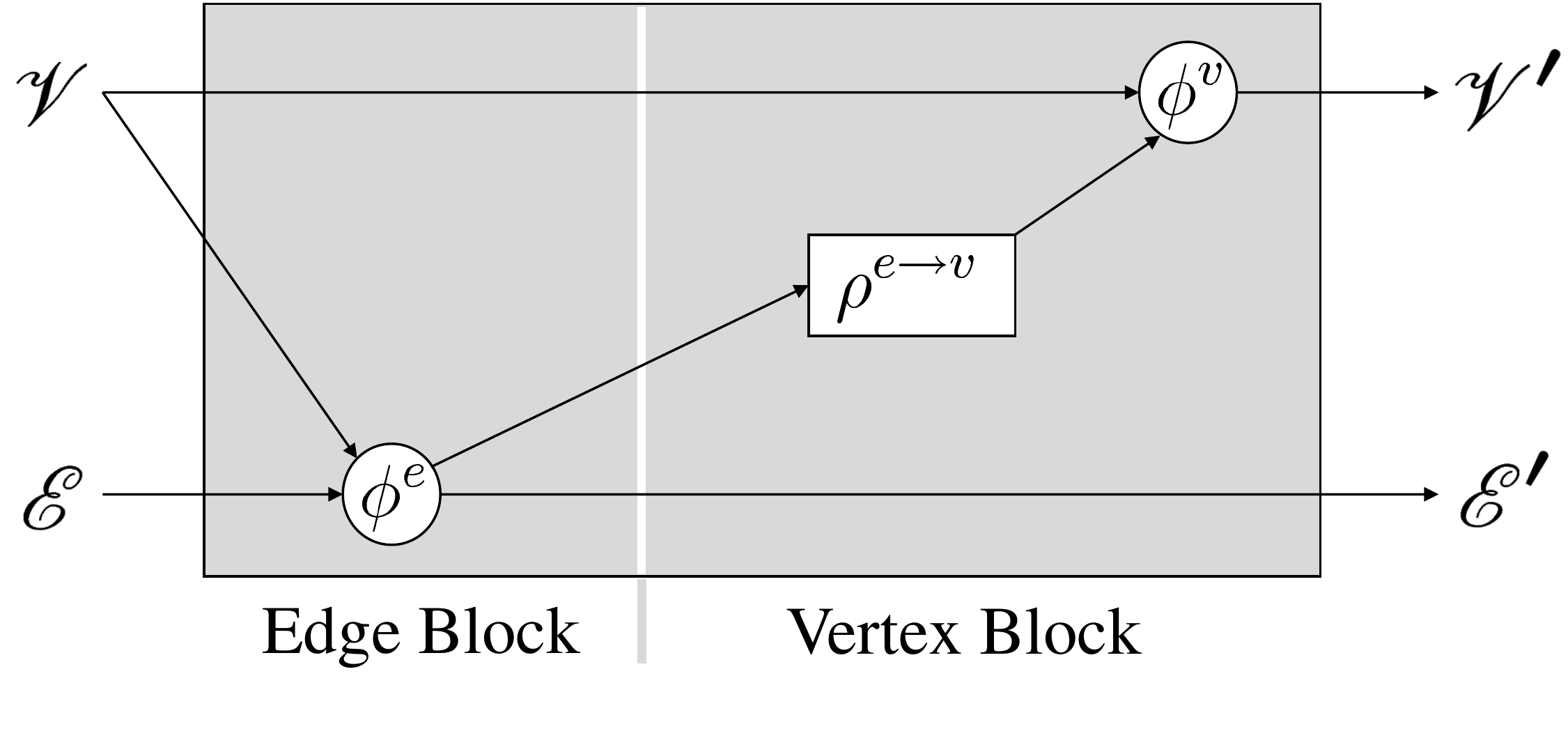}
	\centering
	\caption{An illustration of our proposed simplified GNN block, which only consists of an edge block and a vertex block.}
	\vspace{-1.5em}
	\label{fig.proposed_full_gnn_block}
\end{figure}

According to \cite{fan2018road}, a collection of random variables $\mathscr{X}=\{x_{\mathbf{v}_1}, \dots, x_{\mathbf{v}_{N^{v}}}\}$ depending entirely on their local neighbors in the graph are considered to be in a Markov random field (MRF). Pairwise MRF (pMRF) models are generally used to represent the vertex relations and infer the vertex posterior beliefs \cite{gatterbauer2017linearization}. A pMRF over a graph is associated with a set of vertex potentials as well as edge potentials \cite{koller2009probabilistic}. The overall distribution is the normalized product of all vertex and edge potentials \cite{gatterbauer2017linearization}:
\begin{equation}
P(\mathscr{X})=\frac{1}{Z}
\prod_{\mathbf{v}_i\in\mathscr{V}} \varphi (x_{\mathbf{v}_i}) \prod_{(\mathbf{e},r,s)\in\mathscr{E}}^{} \psi(x_{\mathbf{v}_r},x_{\mathbf{v}_s}),
\end{equation}
where $Z$ is a normalizer, $\varphi$ represents the vertex potential of $\mathbf{v}_i$, and $\psi$ denotes the edge compatibility between the sender $\mathbf{v}_s$ and the receiver $\mathbf{v}_r$. Belief propagation (BP) is commonly used to approximate the posterior belief of a given vertex.  The message $m_{rs}^{(t)}(x_{\mathbf{v}_r})$ sent from $\mathbf{v}_s$ to $\mathbf{v}_r$ in the $t$-th iteration is \cite{gatterbauer2017linearization}:
\begin{equation}
m_{rs}^{(t)}(x_{\mathbf{v}_r})= \varphi(x_{\mathbf{v}_s}) \psi(x_{\mathbf{v}_r},x_{\mathbf{v}_s})  \prod_{k\in\mathscr{N}(\mathbf{v}_s)\setminus {\mathbf{v}_r}}^{}m_{sk}^{(t-1)}(x_{\mathbf{v}_s}),
\label{eq.m}
\end{equation}
where $\mathscr{N}(\mathbf{v}_s)$ is the neighborhood system of $\mathbf{v}_{s}$. The posterior belief of a vertex $x_{\mathbf{v}_i}$ is proportional to the product of the factor and the messages from the variables, namely
\begin{equation}
P^{(t)}(x_{\mathbf{v}_i})\propto \varphi_{i}(x_{\mathbf{v}_i}) \prod_{k\in\mathscr{N}(\mathbf{v}_i)}^{}m_{ik}^{(t)}(x_{\mathbf{v}_i}).
\label{eq.P2}
\end{equation}
It can be found from (\ref{eq.m}) and (\ref{eq.P2}) that the posterior belief of $x_{\mathbf{v}_i}$ is only related to its vertex potential and the edge compatibility between $\mathbf{v}_i$ and its neighbors. Therefore, the global attribute $\mathbf{u}$ in Fig. \ref{fig.full_gnn_block} can be omitted. The simplified GNN block is shown in Fig. \ref{fig.proposed_full_gnn_block}. In this paper, each vertex is considered to have relations with its four closest neighboring vertexes. (\ref{eq.phi^e}) and (\ref{eq.phi^v}) can, therefore, be rewritten as
\begin{equation}
\mathbf{e}'_k = \phi^{e}(\mathbf{e}_k, \mathbf{v}_{r_k}, \mathbf{v}_{s_k}),
\label{eq.phi^e2}
\end{equation}
and
\begin{equation}
\mathbf{v}'_i = \phi^{v}(\bar{\mathbf{e}}_i', \mathbf{v}_{i}),
\label{eq.phi^v2}
\end{equation}
where the multi-layer perceptron is used for $\phi$. Our feature updater then produces an updated vertex feature $\mathbf{R_v}$ of size $(HW) \times C'$ and an updated edge feature $\mathbf{R_e}$ of size $(4HW) \times C'$. These updated features are then processed by an FC layer, a multiplication and a reshaping operator, as shown in Fig. \ref{fig.gal}, to generate an updated tensor (feature representation) $T'$ of size $H \times W \times C'$, which can be considered as a refinement of the input tensor $T$. Considering the balance between the performance improvement and the memory cost, we set the output feature channels to half of the input feature channels, \ie, $C'=\frac{1}{2}C$.


\subsection{GAL-DeepLabv3+}
\label{sec.gal_deeplab}
We implement the developed GAL in {\deeplabvthreeplus} \cite{chen2018encoder} and build a new architecture referred to as GAL-DeepLabv3+, as shown in Fig. \ref{fig.proposed_cnn}. We adopt several residual blocks \cite{he2016deep} as the backbone, and the output feature size from Block5 (the last block) of the adopted backbone is $\frac{H}{16} \times \frac{W}{16} \times C_5$. Then, our GAL takes the feature from Block5 of the backbone as input and outputs the refined feature, which is then concatenated with the input feature and sent to the following ASPP module and the decoder, separately.

\section{Experimental Results and Discussion}
\label{sec.experimental_results_and_discussion}

\begin{figure*}[!t]
	\centering
	\includegraphics[width=1\textwidth]{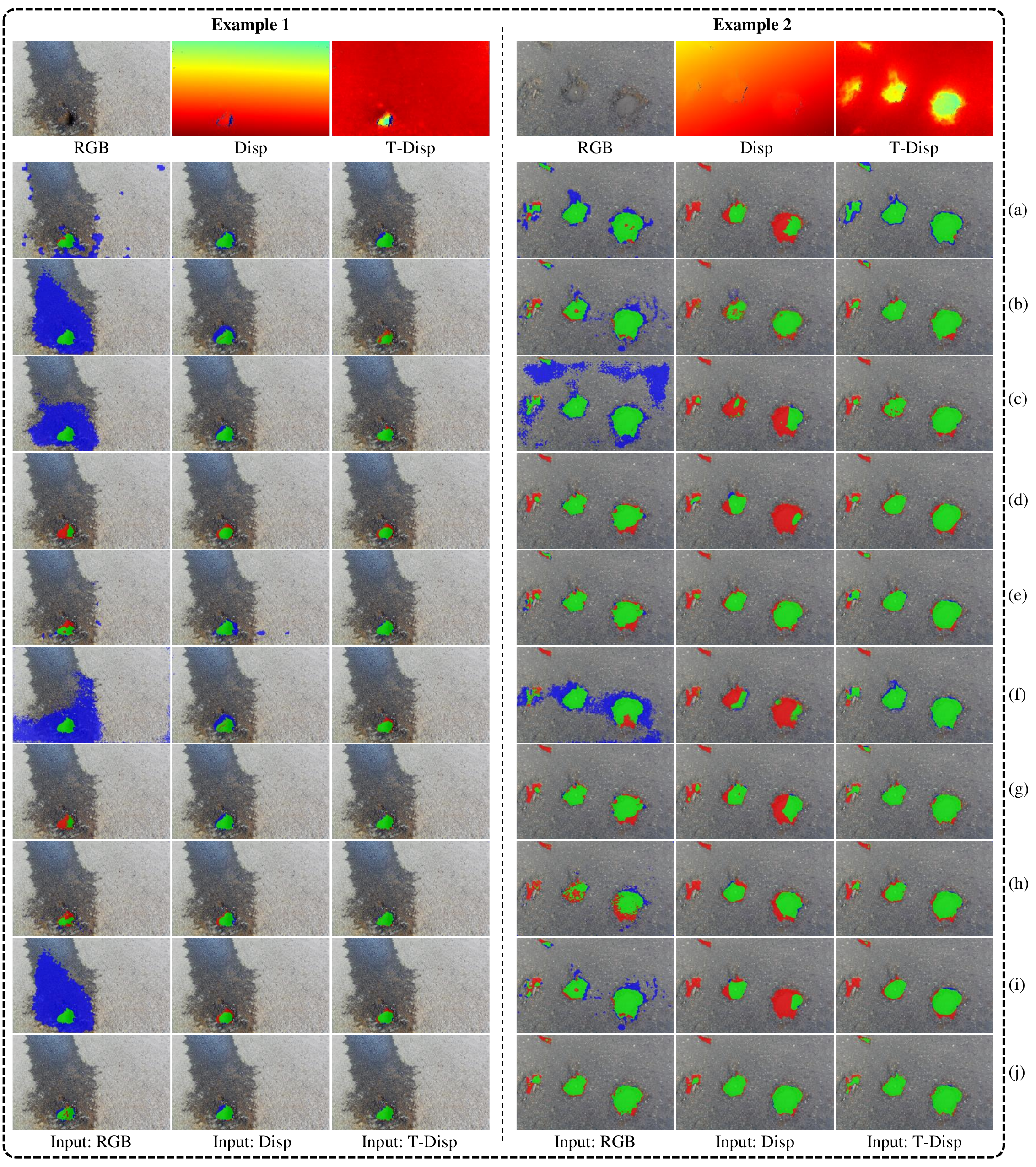}
	\caption{Examples of the experimental results of the nine SoTA CNNs and GAL-DeepLabv3+: (a) FCN \cite{long2015fully}; (b) U-Net \cite{ronneberger2015u}; (c) DenseASPP \cite{yang2018denseaspp}; (d) DUpsampling \cite{tian2019decoders}; (e) GSCNN \cite{takikawa2019gated}; (f) SegNet \cite{badrinarayanan2017segnet}; (g) DeepLabv3+ \cite{chen2018encoder}; (h) PAN \cite{li2018pyramid}; (i) ESPNet \cite{mehta2018espnet}; (j) our developed GAL-DeepLabv3+, where the true-positive, false-positive, and false-negative pixels are shown in green, blue and red, respectively. 
	}
\vspace{-1.5em}
	\label{fig.whole_comparison}
\end{figure*}

\subsection{Dataset and Experimental Setup}
\label{sec.dataset_and_experiment_setup}
We use our recently published road pothole detection dataset \cite{fan2019pothole} to compare the performance of the nine SoTA CNNs mentioned in Section \ref{sec.semantic_segmentation} and our introduced GAL-DeepLabv3+. This dataset contains 55 samples of RGB images (RGB), subpixel disparity images (Disp), and transformed disparity images (T-Disp), which correspond to 14 potholes. The image resolution is $800 \times 1312$ pixels. The 13th and 14th potholes have only one sample each, and therefore they are only used for CNN cross-validation. The remaining 53 samples are divided into 12 sets, which correspond to 12 different potholes. The sample numbers in these 12 sets are: 13, 9, 5, 4, 3, 2, 3, 3, 2, 2, 2, and 5.

In our experiments, we employ the 12-fold cross-validation strategy \cite{geisser1975predictive} to quantify the CNNs' performances, \ie, the performance of each CNN is evaluated 12 times. Each time, a CNN was trained using RGB, Disp, and T-Disp, separately. To quantify the CNNs' performances, we compute the pixel-level precision (Pre), recall (Rec), accuracy (Acc), F-score (Fsc), and intersection over union (IoU). We compute the mean value across the 12 sets for each metric, denoted as mPre, mRec, mAcu, mFsc and mIoU. 

\begin{figure*}[t]
	\centering
	\includegraphics[width=1\textwidth]{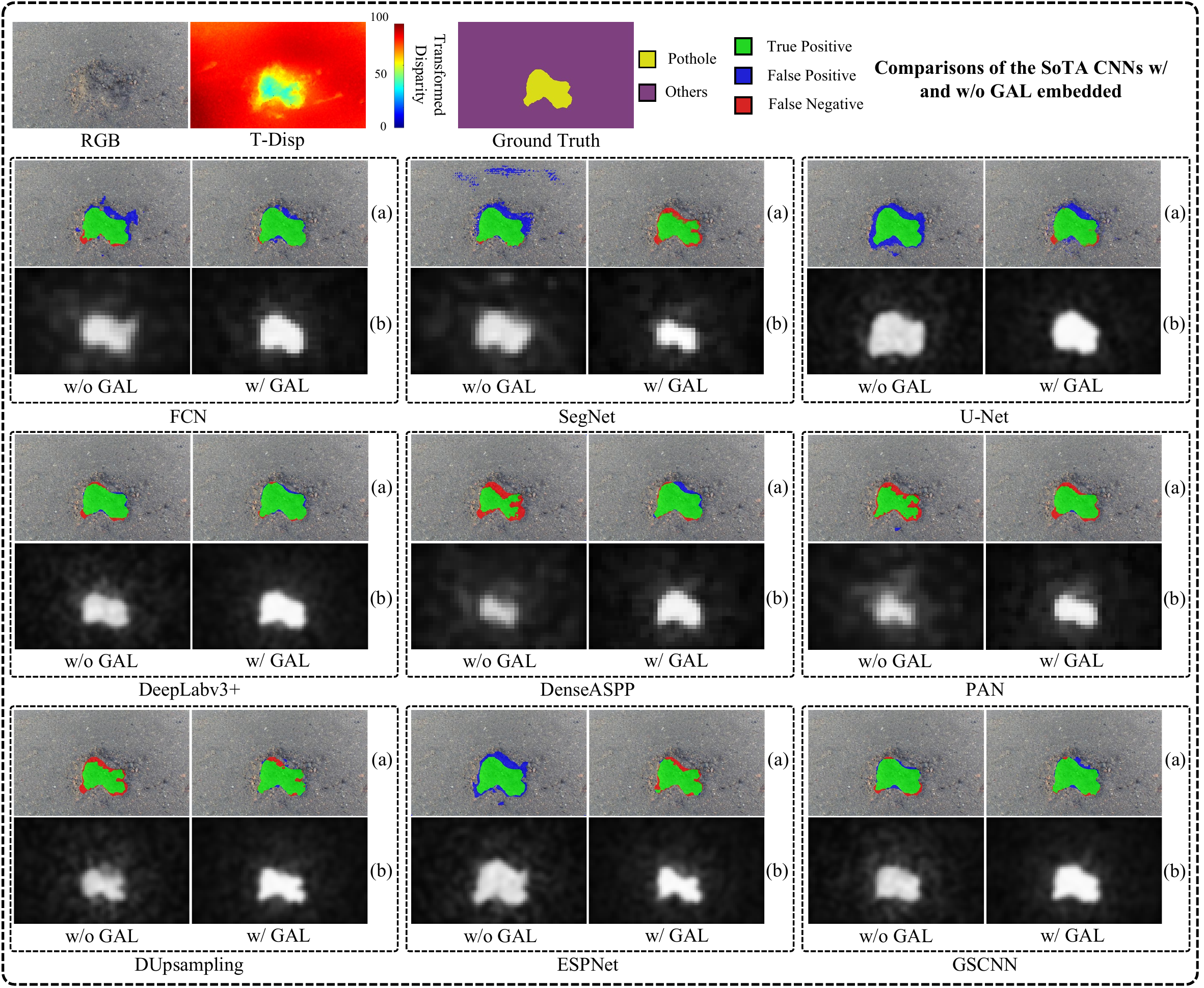}
	\caption{An example of the experimental results of the nine SoTA CNNs without and with GAL embedded: (a) pothole detection results; (b) the corresponding mean activation maps of the features after the last layers of the encoders.}
	\vspace{-1.5em}
	\label{fig.prove_gal}
\end{figure*}

\begin{figure*}[t]
	\centering
	\includegraphics[width=0.98\textwidth]{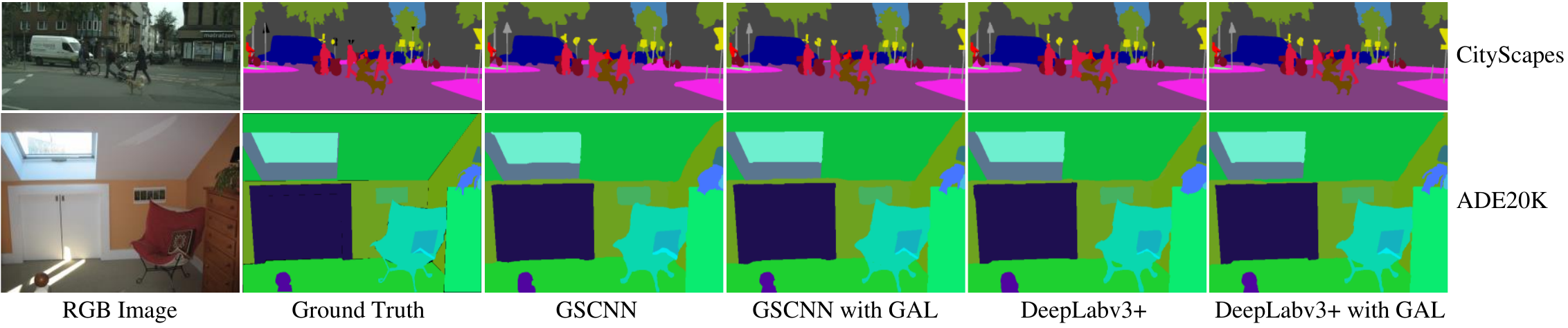}
	\caption{Examples of the experimental results of GSCNN \cite{takikawa2019gated} and DeepLabv3+ \cite{chen2018encoder} with and without GAL embedded.}
	\label{fig.new_datasets}
	\vspace{-0.5em}
\end{figure*}

Additionally, stochastic gradient descent with momentum (SGDM) optimizer \cite{lecun2015deep} is used for CNN training. The maximum epoch for the experiments on RGB, Disp, and T-Disp is set to 150, 100, and 100, respectively. Each network is trained on two NVIDIA GeForce RTX 2080Ti GPUs. We also leverage common training data augmentation techniques, such as random flip, rotation, and translation, to further improve the CNNs' robustness and accuracy.

Next, we conduct ablation studies in Section \ref{sec.ablation_study} to demonstrate the effectiveness of our GAL. Then, a road pothole detection benchmark that provides a detailed performance comparison between the nine SoTA CNNs and our GAL-DeepLabv3+ on the three modalities of training data is presented in Section \ref{sec.benchmark_for_road_pothole_detection}. To further understand how our GAL improves the overall performance for road pothole detection, we implement it in all SoTA CNNs and analyze the feature variation with and without our GAL, as discussed in Section \ref{sec.further_discussion_on_our_gal}.

\begin{table}[t]
	\renewcommand{\arraystretch}{1.3}
	\caption{Ablation studies: (A) shows the baseline results; (B) shows the results of the baseline with GAL embedded; and (C) shows the results of the baseline with ResNet-101 used as the backbone. The best results are shown in bold type.}
	\centering
	\begin{tabular}{lcccc}
		\toprule
		\multicolumn{2}{c}{Setups} & \multicolumn{3}{c}{Evaluation Metrics} \\ \cmidrule(l){1-2} \cmidrule(l){3-5}
		\multicolumn{1}{c}{Backbone} & \multicolumn{1}{c}{GAL} & \multicolumn{1}{c}{mAcc~($\%$)} & \multicolumn{1}{c}{mFsc~($\%$)} & \multicolumn{1}{c}{mIoU~($\%$)} \\ \midrule
		(A) ResNet-50  & -- & 98.453 & 81.479 & 69.011 \\
		(B) ResNet-50  & \cmark & \textbf{98.669} & \textbf{85.636} & \textbf{75.008} \\
		(C) ResNet-101  & -- & 98.581 & 85.167 & 74.228 \\
		\bottomrule
	\end{tabular}
	\vspace{-1.5em}
	\label{tab.ablation_study}
\end{table}

\begin{table*}[!htp]
	\renewcommand{\arraystretch}{1.3}
	\caption{The road pothole detection benchmark conducted with nine SoTA CNNs and our proposed GAL-DeepLabv3+ on three modalities of training data. The best results are shown in bold type.}
	\centering
	\begin{tabular}{L{2.6cm}C{1.6cm}C{1.6cm}C{1.6cm}C{1.6cm}C{1.6cm}C{1.6cm}}
		\toprule
		\multicolumn{1}{c}{Approach} &Input& mPre~($\%$) & mRec~($\%$) & mAcc~($\%$) & mFsc~($\%$) & mIoU~($\%$) \\ \midrule
		\multirow{3}{*}{FCN \cite{long2015fully}} & RGB & 57.029 & 76.324 & 94.881 & 63.670 & 47.647 \\
		& Disp & 85.465 & 49.456 & 97.126 & 60.587 & 43.812 \\
		& T-Disp & 73.970 & 80.874 & 97.386 & 76.940 & 62.698 \\ \midrule
		\multirow{3}{*}{SegNet \cite{badrinarayanan2017segnet}} & RGB & 35.092 & 74.886 & 89.356 & 46.268 & 30.709 \\
		& Disp & 79.007 & 41.299 & 96.578 & 51.603 & 35.211 \\
		& T-Disp & 70.463 & 79.980 & 97.216 & 74.065 & 58.982 \\ \midrule
		\multirow{3}{*}{U-Net \cite{ronneberger2015u}} & RGB & 70.618 & 63.459 & 95.925 & 62.967 & 47.317 \\
		& Disp & 75.872 & 57.253 & 97.062 & 63.886 & 47.205 \\
		& T-Disp & 82.334 & 67.034 & 97.833 & 73.573 & 58.438 \\ \midrule
		\multirow{3}{*}{DeepLabv3+ \cite{chen2018encoder}} & RGB & 78.802 & 66.107 & 97.399 & 70.938 & 55.305 \\
		& Disp & 66.993 & 69.968 & 96.798 & 67.314 & 50.966 \\
		& T-Disp & \textbf{89.819} & 75.154 & 98.453 & 81.479 & 69.011 \\ \midrule
		\multirow{3}{*}{DenseASPP \cite{yang2018denseaspp}} & RGB & 41.818 & 63.212 & 91.113 & 48.929 & 33.191 \\
		& Disp & 68.513 & 63.959 & 96.777 & 64.249 & 47.729 \\
		& T-Disp & 88.914 & 65.637 & 98.018 & 74.398 & 59.824 \\ \midrule
		\multirow{3}{*}{PAN \cite{li2018pyramid}} & RGB & 73.157 & 47.839 & 96.640 & 57.417 & 42.207 \\
		& Disp & 74.522 & 45.762 & 96.352 & 53.290 & 37.146 \\
		& T-Disp & 87.359 & 66.592 & 97.944 & 72.657 & 59.972 \\ \midrule
		\multirow{3}{*}{DUpsampling \cite{tian2019decoders}} & RGB & 75.051 & 66.281 & 97.230 & 69.472 & 53.515 \\
		& Disp & 80.446 & 57.076 & 97.311 & 66.132 & 49.618 \\
		& T-Disp & 86.160 & 72.365 & 98.204 & 77.931 & 64.225 \\ \midrule
		\multirow{3}{*}{ESPNet \cite{mehta2018espnet}} & RGB & 74.547 & 63.694 & 96.133 & 64.850 & 49.327 \\
		& Disp & 85.953 & 51.716 & 97.354 & 63.846 & 47.245 \\
		& T-Disp & 87.190 & 67.644 & 98.038 & 75.358 & 61.013 \\ \midrule
		\multirow{3}{*}{GSCNN \cite{takikawa2019gated}} & RGB & 81.135 & 58.818 & 97.713 & 66.571 & 52.712 \\
		& Disp & 86.275 & 54.612 & 97.407 & 65.481 & 48.992 \\
		& T-Disp & 81.165 & 80.171 & 98.289 & 80.389 & 67.635 \\ \midrule
		\multirow{3}{*}{GAL-DeepLabv3+ (Ours)} & RGB & 83.900 & 69.803 & 97.740 & 74.166 & 60.097 \\
		& Disp & 79.847 & 68.710 & 97.473 & 72.428 & 57.812 \\
		& T-Disp & 89.713 & \textbf{82.205} & \textbf{98.669} & \textbf{85.636} & \textbf{75.008} \\ \bottomrule
	\end{tabular}
	\label{tab.benchmark}
	\vspace{-1.5em}
\end{table*}

\subsection{Ablation Study}
\label{sec.ablation_study}
We adopt DeepLabv3+ \cite{chen2018encoder} as the baseline to conduct ablation studies because it outperforms all other SoTA CNNs. It inputs T-Disp because this modality of vision data is most informative \cite{fan2020we}. Table~\ref{tab.ablation_study} shows the results of the ablation study, where (A) shows the baseline performance, (B) shows the performance of the proposed approach, and (C) shows the performance of the baseline with ResNet-101 as the backbone (abbreviated as ResNet101-DeepLabv3+). Compared to ResNet50-DeepLabv3+, our developed GAL-DeepLabv3+ presents a much better performance. One exciting fact is that our developed GAL-DeepLabv3+ with ResNet-50 as the backbone performs even slightly better than ResNet101-DeepLabv3+. This demonstrates the effectiveness of our proposed GAL.

\subsection{Road Pothole Detection Benchmark}
\label{sec.benchmark_for_road_pothole_detection}
This subsection presents a road pothole detection benchmark with quantitative and qualitative comparisons among nine SoTA CNNs and GAL-DeepLabv3+ trained on the three modalities of vision data. Some examples of the experimental results are shown in Fig. \ref{fig.whole_comparison}. It can be observed that the CNNs trained on RGB can be easily misled by noise, such as a stain on the road (see Fig. \ref{fig.whole_comparison}, Example 1). The CNNs trained on Disp perform slightly better but still produce many false-negative predictions (see Fig. \ref{fig.whole_comparison}, Example 2). By comparison, the CNNs trained on T-Disp perform much more robustly. This is due to that the disparity transformation algorithm makes the damaged road regions become highly distinguishable \cite{fan2019pothole}. Furthermore, Fig.~\ref{fig.whole_comparison} shows that our developed GAL-DeepLabv3+ outperforms all other SoTA CNNs on all three modalities of vision data.

Additionally, the quantitative comparisons are given in Table~\ref{tab.benchmark}, where it can be seen that the mIoU increases by $\sim$11-28$\%$, while the mFsc goes up by $\sim$8-28$\%$ when the CNNs are trained on T-Disp rather than RGB. These results further validate the effectiveness of the disparity transformation algorithm, which converts road disparity information into a more informative format. Furthermore, GAL-DeepLabv3+ outperforms all other SoTA CNNs by $\sim$6-17$\%$ on the mIoU and by $\sim$4-13$\%$ on the mFsc, when trained on T-Disp. This demonstrates that GAL can effectively improve the road pothole detection performance.

\subsection{Further Discussion on GAL}
\label{sec.further_discussion_on_our_gal}
To further understand how GAL improves the CNN's overall performance for road pothole detection, we implement it in each of the nine SoTA CNNs. It should be noted here that we implement GAL after the encoder's last layer for each CNN. The quantitative and qualitative comparisons are given in Table \ref{tab.prove_gal} and Fig. \ref{fig.prove_gal}, respectively. It can be observed that the CNNs with GAL embedded generally perform better than themselves without GAL embedded.

\begin{table}[t]
	\renewcommand{\arraystretch}{1.3}
	\caption{The experimental results of the SoTA CNNs with and without GAL embedded. The best results for each CNN are shown in bold type.}
	\centering
	\begin{tabular}{lC{1.5cm}C{1.5cm}C{1.5cm}}
		\toprule
		\multicolumn{1}{c}{Approach} & mAcc~($\%$) & mFsc~($\%$) & mIoU~($\%$) \\ \midrule
		FCN \cite{long2015fully} & 97.386 & 76.940 & 62.698 \\
		GAL-FCN & \textbf{98.016} & \textbf{80.358} & \textbf{67.205} \\ \midrule
		SegNet \cite{badrinarayanan2017segnet} & 97.216 & 74.065 & 58.982 \\
		GAL-SegNet & \textbf{97.880} & \textbf{79.275} & \textbf{65.703} \\ \midrule
		U--Net \cite{ronneberger2015u} & 97.833 & 73.573 & 58.438 \\
		GAL-U--Net & \textbf{97.958} & \textbf{77.306} & \textbf{63.040} \\ \midrule
		DeepLabv3+ \cite{chen2018encoder} & 98.453 & 81.479 & 69.011 \\
		GAL-DeepLabv3+ & \textbf{98.669} & \textbf{85.636} & \textbf{75.008} \\ \midrule
		DenseASPP \cite{yang2018denseaspp} & 98.018 & 74.398 & 59.824 \\
		GAL-DenseASPP & \textbf{98.177} & \textbf{79.856} & \textbf{66.505} \\ \midrule
		PAN \cite{li2018pyramid} & 97.944 & 72.657 & 59.972 \\
		GAL-PAN & \textbf{98.102} & \textbf{79.194} & \textbf{65.592} \\ \midrule
		DUpsampling \cite{tian2019decoders} & 98.204 & 77.931 & 64.225 \\
		GAL-DUpsampling & \textbf{98.349} & \textbf{81.426} & \textbf{68.713} \\ \midrule
		ESPNet \cite{mehta2018espnet} & 98.038 & 75.358 & 61.013 \\
		GAL-ESPNet & \textbf{98.165} & \textbf{79.747} & \textbf{66.354} \\ \midrule
		GSCNN \cite{takikawa2019gated} & 98.289 & 80.389 & 67.635 \\
		GAL-GSCNN & \textbf{98.486} & \textbf{84.329} & \textbf{72.954} \\ \bottomrule
	\end{tabular}
\vspace{-1.5em}
	\label{tab.prove_gal}
\end{table}

To explore how GAL refines the feature representations, we visualize the mean activation maps of the features output from the encoders' last layers with and without GAL embedded, as shown in Fig. \ref{fig.prove_gal}(b). These maps suggest that GAL can help CNNs concentrate more on the target (road pothole) areas. We believe this is because GAL can be considered as a weight modulation operator, which can effectively augment the activation values in the target areas and reduce the activation values in the background areas. A typical convolutional layer can be formulated as follows:
\begin{equation}
    y(x)=\sum_{n=1}^{N}w_n \cdot x(n).
    \label{eq.conv1}
\end{equation}
Now, with GAL embedded, we can formulate this process as follows:
\begin{equation}
    y(x)=\sum_{n=1}^{N}w_n  \cdot \Big( \bar{x}(n) \cdot \Delta w_e \Big ),
    \label{eq.conv2}
\end{equation}
where $\Delta w_e$ is obtained from the updated edge features, and $\bar{x}(n)$ is the updated vertex features from $x(n)$. Based on (\ref{eq.conv1}) and (\ref{eq.conv2}), we can conclude that GAL can be considered as an effective and efficient weight modulation operator, which can greatly refine the feature representations, thus improving the CNN's overall performance for road pothole detection.

\begin{table}[t]
	\renewcommand{\arraystretch}{1.3}
	\caption{The Experimental Results of GSCNN \cite{takikawa2019gated} and DeepLabv3+ \cite{chen2018encoder} with and without our GAL embedded on the Cityscapes \cite{cordts2016cityscapes} and ADE20K \cite{zhou2017scene} Datasets. The best results for each CNN are shown in bold type.}
	\centering
	\begin{tabular}{L{2.2cm}C{1.1cm}C{1.1cm}C{1.1cm}C{1.1cm}}
		\toprule
		\multicolumn{1}{c}{\multirow{2}{*}{Approach}} & \multicolumn{2}{c}{Cityscapes \cite{cordts2016cityscapes}} & \multicolumn{2}{c}{ADE20K \cite{zhou2017scene}} \\ \cmidrule(l){2-3} \cmidrule(l){4-5}
		&                                        mFsc~(\%)                            & mIoU~(\%)                            & mFsc~(\%)                                                  & mIoU~(\%)                 \\ \midrule
		GSCNN \cite{takikawa2019gated} & 86.383 & 76.829 & 59.727 & 42.556 \\
		GAL-GSCNN & \textbf{88.920} & \textbf{80.186} & \textbf{62.962} & \textbf{45.934} \\ \midrule
		DeepLabv3+ \cite{chen2018encoder} & 87.198 & 77.398 & 60.685 & 43.528 \\
		GAL-DeepLabv3+ & \textbf{89.802} & \textbf{81.537} & \textbf{64.120} & \textbf{47.291} \\ \bottomrule
	\end{tabular}
	\label{tab.new_datasets}
	\vspace{-1.5em}
\end{table}

\section{Discussion}
\label{sec.discussion}
Potholes are a common type of road distress. The detection of other categories of road distresses, such as cracks, typically requires different kinds of computer vision algorithms. For example, the SoTA road crack detection algorithms \cite{zhang2016road, oliveira2012automatic, shi2016automatic} commonly leverage image classification CNNs instead of semantic segmentation CNNs to identify whether an image patch contains cracks because road cracks cannot be easily identified from depth/disparity images, and the semantic segmentation CNN is challenging to retain highly accurate semantic content for such tiny objects. Furthermore, although different types of road distress can be recognized and classified with SoTA object detection algorithms, such as YOLO-v3 \cite{redmon2018yolov3} utilized in \cite{du2020pavement}, such road distress detection results can only be at instance level instead of pixel level. The measurement of a road pothole's volume typically requires pixel-level predictions instead of a region of interest. 

In addition to road pothole detection, our introduced GAL can also be embedded in CNNs to solve other semantic image segmentation problems. To demonstrate its feasibility in other challenging multi-class scene understanding applications, we train GSCNN \cite{takikawa2019gated} and DeepLabv3+ \cite{chen2018encoder} both with and without GAL embedded on the Cityscapes \cite{cordts2016cityscapes} and ADE20K \cite{zhou2017scene} datasets. The Cityscapes dataset \cite{cordts2016cityscapes} was created for semantic urban scene understanding, while the ADE20K dataset \cite{zhou2017scene} (including diverse scenarios) was created for general scene parsing.
The qualitative and quantitative results are given in Fig. \ref{fig.new_datasets} and Table \ref{tab.new_datasets}, respectively. It can be observed that both CNNs with GAL embedded can produce more accurate results, where the mIoU increases by $\sim$4\% and the mFsc increases by $\sim$3\%. These results suggest the generalizability of our proposed GAL for other challenging semantic segmentation tasks. Therefore, we believe our proposed GAL can be easily incorporated into any existing CNN to achieve SoTA semantic scene understanding performance.

\section{Conclusion and Future Work}
\label{sec.conclusion_future_work}
In this paper, we provided a comprehensive study on road pothole detection including building up a benchmark, developing a novel layer based on GNN, and proposing an effective and efficient CNN for road pothole detection. Experiments verify that our proposed GAL can effectively refine the feature representations and thus improve the overall semantic segmentation performance. Moreover, the transformed disparity images can make road potholes highly distinguishable and benefit all CNNs for road pothole detection. Compared with the SoTA CNNs, our proposed GAL-DeepLabv3+ achieves superior performance and produces more robust and accurate results. We believe that the provided benchmark and our proposed models are helpful to stimulate further research in this area. Furthermore, our proposed techniques can also be employed to solve other general semantic segmentation/understanding problems. In the future, we will continue to explore graph-based architectures that can optimize the feature representations effectively and efficiently for semantic segmentation.

\bibliographystyle{IEEEtran}

\end{document}